\documentclass[draft=false,11pt,a4paper]{article}
\PassOptionsToPackage{breaklinks,final}{hyperref}
\usepackage{emnlp-ijcnlp-2019}
\usepackage{times}
\usepackage{fdsymbol}
\usepackage{latexsym}
\usepackage{url}
\aclfinalcopy

\usepackage{booktabs}
\usepackage{pgf}
\usepackage{subcaption}
\usepackage[textsize=small]{todonotes}

\newcommand{\componentletter}[1]{\textbf{\underline{#1}}}

\title{Few-Shot and Zero-Shot Learning for Historical Text Normalization}
\author{Marcel Bollmann$^\clubsuit$ \and Natalia Korchagina$^\diamondsuit$ \and Anders S{\o}gaard$^\clubsuit$\\[.3em]
  $^\clubsuit$Department of Computer Science, University of Copenhagen\\
  $^\diamondsuit$Institute of Computational Linguistics, University of Zurich\\[.3em]
  \texttt{marcel@di.ku.dk, korchagina@ifi.uzh.ch, soegaard@di.ku.dk}
 }

\begin{document}
\maketitle
\begin{abstract}
Historical text normalization often relies on small training datasets. Recent work has shown that multi-task learning can lead to significant improvements by exploiting synergies with related datasets, but there has been no systematic study of different multi-task learning architectures. This paper evaluates 63~multi-task learning configurations for sequence-to-sequence-based historical text normalization across ten datasets from eight languages, using autoencoding, grapheme-to-phoneme mapping, and lemmatization as auxiliary tasks. We observe consistent, significant improvements across languages when training data for the target task is limited, but minimal or no improvements when training data is abundant. We also show that zero-shot learning outperforms the simple, but relatively strong, identity baseline. 
\end{abstract}

\section{Introduction}

Historical text normalization is the task of mapping variant spellings in
historical documents---e.g., digitized medieval manuscripts---to a common form,
typically their modern equivalent.  The aim is to make these documents amenable
to search by today's scholars, processable by NLP~tools, and accessible to lay
people.  Many historical documents were written in the absence of standard
spelling conventions, and annotated datasets are rare and small, making
automatic normalization a challenging
task~\citep[cf.][]{Piotrowski2012,BollmannPhD}.

In this paper, we experiment with datasets in eight different languages:
English, German, Hungarian, Icelandic, Portuguese, Slovene, Spanish, and
Swedish.  We use a standard neural sequence-to-sequence model, which has been
shown to be competitive for this
task~\citep[e.g.,][]{Korchagina2017,BollmannPhD,Tang-etal2018}.  Our main focus
is on analyzing the usefulness of multi-task learning strategies (a)~to leverage
whatever supervision is available for the language in question ({\em few-shot
  learning}), or (b)~to {do away with} the need for supervision in the target
language altogether ({\em zero-shot learning}).

\newcite{Bollmann-etal2017} previously showed that multi-task learning with grapheme-to-phoneme conversion as an auxiliary task improves a sequence-to-sequence model for historical text normalization of German texts; \newcite{Bollmann-etal2018} showed that multi-task learning is particularly helpful in low-resource scenarios.  We consider three auxiliary tasks in our experiments---grapheme-to-phoneme mapping, autoencoding, and lemmatization---and focus on extremely low-resource settings.

Our paper makes several contributions:
\begin{enumerate}
\item[(a)] We evaluate 63 multi-task learning configurations across ten datasets in eight languages, and with three different auxiliary tasks. 
\item[(b)] We show that in few-shot learning scenarios (ca.\,1,000~tokens), multi-task learning leads to robust, significant gains over a state-of-the-art, single-task baseline.\footnote{We note that 1,000 tokens is more instances than is typically considered in few-shot learning; e.g., \newcite{Kimura:ea:18} use up to 200 instances. We argue that for structured prediction it is reasonable to assume more data, yet we also consider scenarios down to as little as 100~instances.}
\item[(c)] We are, to the best of our knowledge, the first to consider {\em zero-shot historical text normalization}, and we show significant improvements over the simple, but relatively strong, identity baseline.
\end{enumerate}

While our focus is on the specific task of historical text normalization, we believe that our results can be of interest to anyone looking to apply multi-task learning in low-resource scenarios.

\begin{table}
  \centering
  \begin{tabular}{llr}
\toprule
\multicolumn{2}{l}{\textbf{Dataset/Language}} & \textbf{Tokens (Dev)} \\
\midrule
DE\textsubscript{A} & German (Anselm) & 45,996 \\
DE\textsubscript{R} &  German (RIDGES) & 9,712 \\
EN & English & 16,334 \\
ES & Spanish & 11,650 \\
HU & Hungarian & 16,707 \\
IS & Icelandic & 6,109 \\
PT & Portuguese & 26,749 \\  
SL\textsubscript{B} & Slovene (Bohori{\v c}) & 5,841 \\ 
SL\textsubscript{G} & Slovene (Gaj) & 20,878 \\
SV & Swedish & 2,245 \\
 \bottomrule
\end{tabular}
  \caption{Historical datasets used in our experiments and the size of their development sets.  (Size of the training sets is fixed in all our experiments.)}
  \label{tab:datasets}
\end{table}

\paragraph{Datasets}
We consider ten datasets spanning eight languages, taken from \citet{Bollmann2019}.\footnote{The datasets are available from:\\\url{https://github.com/coastalcph/histnorm}}
Table~\ref{tab:datasets} gives an overview of the languages and the size of the development set, which we use for evaluation.

\section{Model architecture}
\label{sec:architecture}

We use a standard attentional encoder--decoder architecture~\citep{Bahdanau-etal2014} with words as input sequences and characters as input symbols.
\footnote{Our implementation uses the XNMT~toolkit~\citep[\url{https://github.com/neulab/xnmt}]{Neubig-etal2018}.}
Following the majority of previous work on this topic (cf.\ Sec.\,\ref{sec:related-work}), we limit ourselves to word-by-word normalization, ignoring problems of contextual ambiguity.
Our model consists of the following parts (which we will also refer to using the bolded letters):

\begin{itemize}
  \item \componentletter{S}ource embedding layer: transforms input characters into dense vectors.
  \item \componentletter{E}ncoder: a single bidirectional LSTM that encodes the embedded input sequence.
  \item \componentletter{A}ttention layer: calculates attention from the encoded inputs and the current decoder state using a multi-layer perceptron~\citep[as in][]{Bahdanau-etal2014}.
  \item \componentletter{T}arget embedding layer: transforms output characters into dense vectors.
  \item \componentletter{D}ecoder: a single LSTM that decodes the encoded sequence one character at a time, using the attention vector and the embedded previous output characters as input.
  \item \componentletter{P}rediction layer: a final feed-forward layer that linearly transforms the decoder output and performs a softmax to predict a distribution over all possible output characters.
\end{itemize}

\paragraph{Hyperparameters}  We tuned our hyperparameters on the English development section. We use randomly initialized embeddings of dimensionality~60, hidden layers of dimensionality~300, a dropout of~0.2 and a batch size of~30.  We train the model for an unspecified number of epochs, instead relying on early stopping on a held-out validation set.  Since we experiment with varying amounts of training data, we choose to derive this held-out data from the given training set, using only 90\%~of the tokens as actual training data and the remaining~10\% to determine early stopping.

\section{Multi-task learning}
\label{sec:mtl}

Multi-task learning~(MTL) is a technique to improve generalization by training a model jointly on a set of related tasks.  We follow the common approach of hard parameter sharing suggested by~\citet{Caruana1993}, in which certain parts of a model architecture are shared across all tasks, while others are kept distinct for each one.  Such approaches have been applied successfully to a variety of problems, e.g., machine translation~\cite{Dong-etal2015}, sequence labelling~\cite{Yang-etal2016,Peng-Dredze2017}, or discourse parsing~\cite{Braud-etal2016}. 

\paragraph{Auxiliary tasks}
We experiment with the following auxiliary tasks:

\begin{itemize}
\item \textbf{Autoencoding.}  We use data extracted from Wikipedia\footnote{Whenever possible, we used the dumps provided by the Polyglot project: \url{https://sites.google.com/site/rmyeid/projects/polyglot}\\Since an Icelandic text dump was not available from Polyglot, we generated one ourselves using the Cirrus Extractor: \url{https://github.com/attardi/wikiextractor}\\All dumps were cleaned from punctuation marks.} and train our model to recreate the input words.  In the normalization task, large parts of the input words often stay the same, so autoencoding might help to reinforce this behavior in the model.
\item \textbf{Grapheme-to-phoneme mapping (g2p).}  This task uses the data by~\citet{Deri-Knight2016} to map words (i.e., sequences of graphemes) to sequences of phonemes.  \citet{Bollmann-etal2017} previously showed that this task can improve historical normalization, possibly because changes in spelling are often motivated by phonological processes, an assumption also made by other normalization systems~\citep{Porta-etal2013,Etxeberria-etal2016}.
\item \textbf{Lemmatization.}  We use the UniMorph dataset~\citep{UniMorph}\footnote{\url{https://unimorph.github.io/}} to learn mappings from inflected word forms to their lemmas.  This task is similar to normalization in that it maps a set of different word forms to a single target form, which typically bears a high resemblance to the input words.
\end{itemize}
Since we train separate models for each historical dataset, we always use auxiliary data from the same language as the dataset.

\paragraph{Training details}\label{par:training-details}
When training an MTL~model, we make sure that each training update is based on a balanced combination of main and auxiliary task inputs; i.e., for each batch of 30~tokens of the historical normalization task, the model will see 10~tokens from each auxiliary task.  Epochs are still counted based on the normalization task only.  This way, we try to make up for the imbalanced quantity of different auxiliary datasets.

\begin{figure}[t]
    \centering
    \input{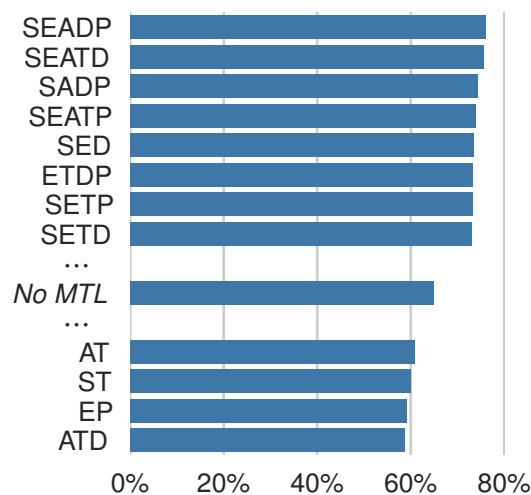}
    \caption{Normalization accuracy on the English-1k dataset, trained jointly with all three auxiliary tasks; letters indicate which model components (cf.\ Sec.\,\ref{sec:architecture}) are shared between tasks.}
    \label{fig:mtl-combinations}
\end{figure}

\begin{figure}[t]
    \hspace{-0.8em}
    \input{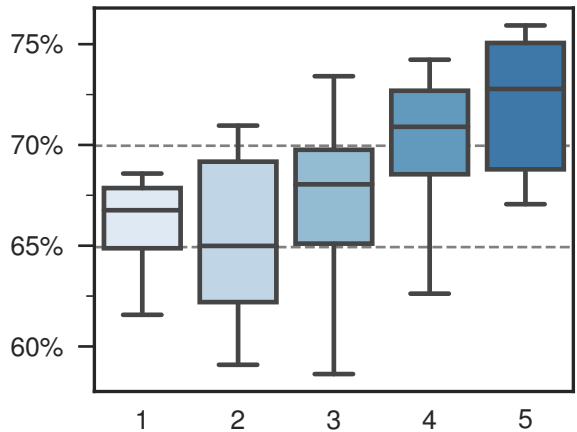}
    \caption{Quartiles of the normalization accuracies (on English-1k) by the number of shared components in the MTL~model; bottom dashed line indicates no shared components (=~single-task), top dashed line indicates all (=~6) shared components.}
    \label{fig:mtl-num-shared}
\end{figure}

\subsection{Experiment~1: What to share?}
\label{sec:what-to-share}

In previous work on multi-task learning, there is no clear consensus on which parts of a model to share and which to keep separate.  \citet{Bollmann-etal2017} share all parts of the model except for the final prediction layer, while other multi-task sequence-to-sequence models keep task-specific encoders and decoders (cf.\ also Sec.\,\ref{sec:related-work}).  In principle, though, the decision to share parameters between tasks can be made for each of the encoder--decoder components individually, allowing for many more possible MTL~configurations.

\paragraph{Setup}
We explore the effect of different sharing configurations.
The architecture described in Sec.~2 leaves us with $2^6 = 64$ possible model configurations. When all parameters are shared, this is identical to training a single model to perform all tasks at once; when none are shared, this is identical to a single-task model trained on historical normalization only. We identify an MTL configuration using letters (cf.\ the bold letters from Sec.\,\ref{sec:architecture}) to indicate which parts of the model are shared; e.g., an ``SE''~model would share the source embeddings and the encoder, an ``SEATD''~model would share everything except the final prediction layer, and so on.

In Experiment 1, we only use the first 1,000~tokens of the English historical dataset for training.  We combine this with all three auxiliary tasks (using their full datasets) and train one MTL~model for each of the 64~different sharing configurations.

\paragraph{Results}
Figure~\ref{fig:mtl-combinations} shows an excerpt of the results, evaluated on the dev~set of the English dataset.  The best MTL~model achieves a normalization accuracy of~75.9\%, while the worst model gets~58.6\%.  In total, 49~configurations outperform the single-task model, showing the general effectiveness of the MTL~approach. Sharing more is generally better; nine out of the top ten configurations share at least four components.  Figure~\ref{fig:mtl-num-shared} visualizes the accuracy distribution by the number of shared components in the MTL~model, supporting this conclusion.

\begin{figure}[t]
    \hspace{-0.8em}
    \input{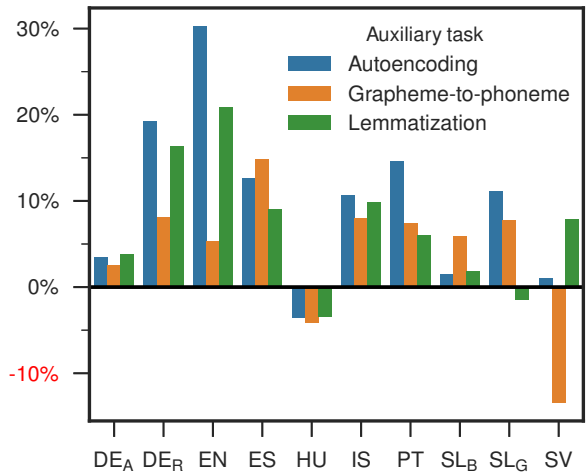}
    \caption{Error reduction for the SEADP configuration by auxiliary task, using 1,000~tokens from the historical datasets for training.}
    \label{fig:mtl-aux}
\end{figure}

\subsection{Experiment~2: Which auxiliary tasks?}
\label{sec:what-aux-task}

In the previous experiment, we trained the models using all three auxiliary tasks at the same time.  However, not all of these tasks might be equally helpful for learning the normalization task.  While \citet{Bollmann-etal2017} show the effectiveness of the grapheme-to-phoneme task, they only evaluate on German, and autoencoding and lemmatization have so far not been evaluated at all for improving historical text normalization.

\paragraph{Setup}
We want to investigate the improvements from each auxiliary task in isolation compared to (a)~the single-task baseline and (b)~the previous approach of training with all three auxiliary tasks simultaneously.  For this, we select the best MTL~configuration from Sec.\,\ref{sec:what-to-share}, which is to share everything except the target embeddings~(``SEADP''), and train one single-task model and four MTL~models per dataset: one for each of the three auxiliary tasks, and one that uses all three tasks at the same time.

As before, we only use the first 1,000~tokens of each historical dataset.  This also makes the results more comparable across datasets, as the size of the training set for the main task can affect the usefulness of multi-task learning.\footnote{The same is true, of course, for the size of the auxiliary datasets.  We try to balance out this factor by balancing the training updates as described in Sec.\,\ref{par:training-details}~``Training details'', but we also note that we do not observe a correlation between auxiliary dataset size and its effectiveness for~MTL in Fig.\,\ref{fig:mtl-aux}.}

\makeatletter
\newlength\mysuperscriptlen
\DeclareRobustCommand*\Textsuperscript[1]{%
\@Textsuperscript{\selectfont#1}}
\def\@Textsuperscript#1{%
\settoheight\mysuperscriptlen{\fontsize\f@size\z@ A}%
{\m@th\ensuremath{\raise.3\mysuperscriptlen\hbox{\fontsize\sf@size\z@#1}}}}
\makeatother
\newcommand{\notsignificant}[0]{}

\begin{table*}
\begin{subtable}[t]{0.53\textwidth}
  \centering
  \begin{small}
  \begin{tabular}{lrrrrr}
\toprule
\textbf{Dataset} & {\textbf{Single}} & \multicolumn{4}{c}{\textbf{Multi-task}} \\
\cmidrule(lr){3-6}
& & Autoenc & Lemma & g2p & \textsc{All~3} \\
\midrule
DE\textsubscript{A} & 54.84 & 56.41 & \textbf{56.55} & 55.99 & 56.52 \\
DE\textsubscript{R} & 56.72 & \textbf{65.05} & 63.79 & 60.25 & 64.49 \\
EN & 66.95 & \textbf{76.94} & 73.84 & 68.72 & 72.01 \\
ES & 74.68 & 77.87 & 76.97 & 78.45 & \textbf{79.09} \\
HU & \textbf{42.44} & 40.39 & 40.49 & 40.07 & 38.64 \\
IS & 63.40 & 67.31 & 67.02 & 66.31 & \textbf{68.51} \\
PT & 72.23 & \textbf{76.28} & 73.89 & 74.27 & 75.55 \\  
SL\textsubscript{B} & 74.06 & \notsignificant{74.44} & \notsignificant{74.54} & \textbf{75.59} & \notsignificant{74.39} \\ 
SL\textsubscript{G} & 86.34 & 87.86 & \notsignificant{86.15} & 87.40 & \textbf{89.45} \\
SV & 69.98 & \notsignificant{70.29} & 72.34 & 65.97 & \textbf{73.05} \\
\midrule
Micro-Avg & 64.46 & \textbf{67.46} & 66.47 & 65.79 & 67.04 \\
 \bottomrule
\end{tabular}
\end{small}
  \caption{Single-task vs.\ multi-task models}
  \label{tab:results-mtl}
\end{subtable}
\hspace{.2em}
\begin{subtable}[t]{0.43\textwidth}
  \centering
 \begin{small}
 \begin{tabular}{lrrrr}
\toprule
\textbf{Dataset} & \textbf{Best in~(a)} & \multicolumn{3}{c}{\textbf{from \citet{Bollmann2019}}} \\
\cmidrule(lr){3-5}
&& Norma & SMT & NMT 
\\
\midrule
DE\textsubscript{A} & 56.55 & \textbf{61.27} & 58.60 & 52.74 \\
DE\textsubscript{R} & 65.05 & 73.62 & \textbf{75.04} & 60.61 \\
EN & 76.94 & \textbf{84.53} & 83.81 & 66.93 \\
ES & 79.09 & \textbf{86.21} & 85.89 & 76.32 \\
HU & 42.44 & \textbf{55.75} & 53.00 & 40.52 \\
IS & 68.51 & 70.86 & \textbf{72.30} & 62.80 \\
PT & 76.28 & \textbf{82.94} & 82.00 & 71.43 \\  
SL\textsubscript{B} & 75.59 & 78.97 & \textbf{82.90} & 73.83 \\ 
SL\textsubscript{G} & 89.45 & 84.36 & \textbf{90.00} & 86.31 \\
SV & 73.05 & 74.54 & \textbf{78.51} & 66.43 \\
\midrule
Micro-Avg & 68.13 & \textbf{73.30} & 73.07 & 63.80 \\
 \bottomrule
\end{tabular}
\end{small}
  \caption{Comparison to previous work}
  \label{tab:results-previous}
\end{subtable}
  \caption{Normalization accuracy on dev sets after training on 1,000~tokens.  Best results highlighted in bold.
  }
\end{table*}

\paragraph{Results}
Figure~\ref{fig:mtl-aux} shows the error reduction of the MTL~models compared to the single-task setup.  For most datasets, MTL improves the results; the main exception is Hungarian, where all three auxiliary tasks lead to a loss in accuracy. The results show that not all auxiliary tasks are equally beneficial.  Autoencoding provides the largest error reduction in most cases, while lemmatization is often slightly worse, but provides the best result for German~(Anselm) and Swedish.  The grapheme-to-phoneme task, on the other hand, performs worst on average, yielding much less benefits on German~(Ridges) and English, and even \textit{increases} the error on Swedish.

Table~\ref{tab:results-mtl} shows the accuracy scores for all datasets and models.  The full MTL~model---training jointly on all tasks---only achieves the best performance on four of the datasets.  Since the dev sets used for this evaluation vary strongly in size, we also calculate the {\em micro-average} of the accuracy scores, i.e., the accuracy obtained over the concatenation of all datasets.  Here, we can see that using only autoencoding as an auxiliary task actually produces the highest average accuracy.

\subsection{Experiment~3: How much training data?}
\label{sec:learning-curves}

All previous experiments have used 1,000~tokens from each historical dataset for training.
\citet{Bollmann-etal2018} show that the benefits of multi-task learning depend on training data quantity, so it is unclear whether the findings generalize to smaller or larger datasets.

\paragraph{Setup}
We analyze the benefit of MTL depending on the amount of training data that is used for the main task.  We do this by training MTL~models (using all three auxiliary tasks, as in Sec.\,\ref{sec:what-to-share}) with varying amounts of historical training data, ranging from 100~tokens to 50,000~tokens.  Different sharing configurations might conceivably give different benefits based on the training set size.  We therefore evaluate each of the top three MTL~configurations from Sec.\,\ref{sec:what-to-share}, as well as the single-task model, across different data sizes.

\begin{figure*}[p] 
    \centering
    \hspace{-0.5em}
    \input{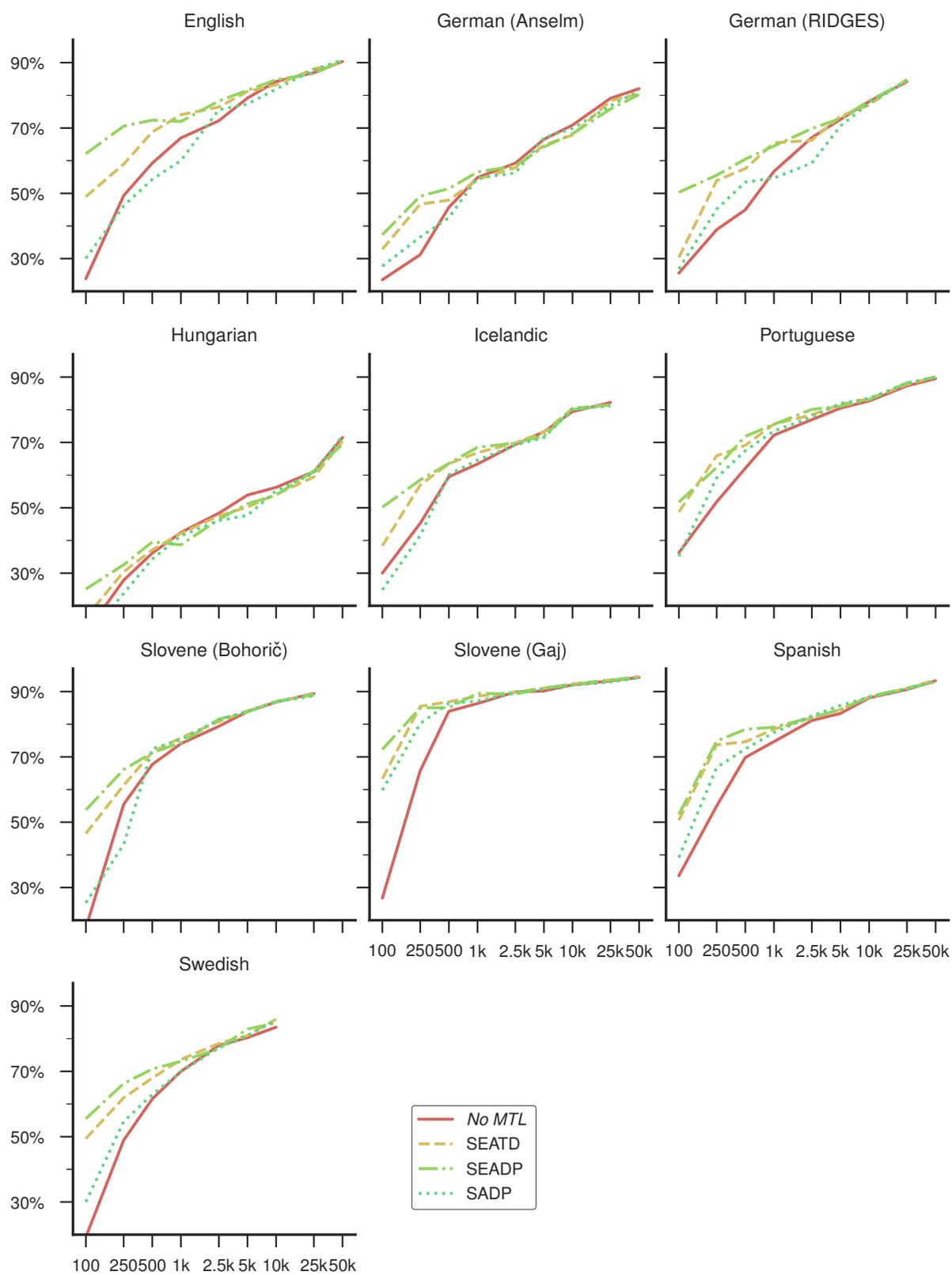}
    \caption{Learning curves for all datasets, showing the normalization accuracy of a single-task and three multi-task learning models in relation to the training set size; note that the $x$-axis is log-scaled.}
    \label{fig:learning-curves}
\end{figure*}

\paragraph{Results}
Figure~\ref{fig:learning-curves} shows learning curves for all of our historical datasets.  The quantity of improvements from MTL differs between datasets, but there is a clear tendency for MTL to become less beneficial as the size of the normalization training set increases.  In some cases, using MTL with larger training set sizes even results in \emph{lower} accuracy compared to training a single-task model to do normalization only.  This suggests that multi-task learning---at least with the auxiliary tasks we have chosen here---is mostly useful when the training data for the main task is sparse.

\begin{figure}[t]
    \hspace{-0.8em}
    \input{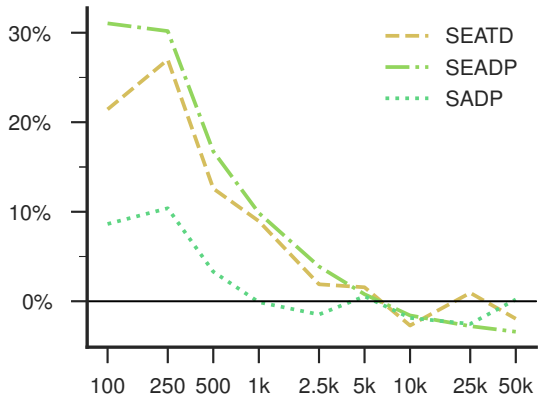}
    \caption{Error reduction for three MTL~configurations by training set size, (micro-)averaged over all datasets.}
    \label{fig:mtl-diff}
\end{figure}

Since the accuracy scores of the different models are often within close range of each other, Figure~\ref{fig:mtl-diff} visualizes the three MTL~configurations in terms of error reduction compared to the single-task model, averaged over all ten datasets.  This again highlights the decreasing gains from MTL with increasing amounts of training data.

\subsection{Comparison to previous work}
\label{sec:comparison}

\citet{Bollmann2019} compares normalization models when trained with different amounts of data, including a setting with 1,000~tokens for training, allowing us to directly compare our results with those reported there.\footnote{\citet{Bollmann2019} only shows graphical plots for these results, but the exact figures were released at: \url{https://github.com/coastalcph/histnorm/blob/master/appendix_tab6.pdf}}
These results are shown in Table~\ref{tab:results-previous}.
Comparing our single-task system with their NMT~model (which is very similar to ours), we see that the scores are overall comparable, suggesting that our implementation is sound.  At the same time, our best scores with MTL are still far below those produced by SMT or the rule-based ``Norma''~tool.  This, unfortunately, is a negative result for the neural approach in this low-resource scenario, and the diminishing gains from MTL that were shown in Sec.\,\ref{sec:learning-curves} suggest that our presented approach will not be sufficient for elevating the neural model above its non-neural alternatives for this particular task.

\subsection{Experiment~4: Zero-shot learning}
\label{sec:zeroshot}

Most previous work on historical text normalization has focused on a supervised scenario where some labeled data is available for the target domain, i.e., the particular historical language you are interested in. 
Since spelling variation is highly idiosyncratic in the absence of normative spelling guidelines, models are not expected to generalize beyond specific language stages, or sometimes even manuscript collections. This means that many historical text normalization projects require resources to annotate new data. This paper is the first to experiment with a zero-shot learning scenario that leverages existing data from other languages, but assumes {\em no}~labeled data for the target language.

\paragraph{Setup}
For the zero-shot experiments, we use the same model as for the single-task baseline; in other words, all layers are shared between all tasks and languages.  Instead, to allow the model to discern between languages and tasks, we prepend two extra symbols to all model inputs: a \emph{language identifier} and a \emph{task identifier.} For each language, we then train a single model on all tasks---normalization, lemmatization, autoencoding, and grapheme-to-phoneme transduction---and all languages, \emph{except} for the normalization task of the target language.  This way, the model can observe data from the normalization task (albeit in other languages) and from the target language (albeit from auxiliary tasks only), but does not see any normalization data from the target language.  In those cases where there are two datasets from the same language, we leave out {\em both} of them from the training step.  The model is similar to previous work on zero-shot neural machine translation~\cite{Johnson-etal2016}.

As before, we include only 1,000~tokens from each historical dataset for training.  In each training update, we use an equal number of samples from each dataset/task combination, and define an epoch to consist of 1,000~samples from each of these combinations.  Since we do not want to feed the model any normalization data from the target language during training, we cannot use early stopping, but instead train for a fixed number of 10~epochs.

\begin{table}
  \centering
 \begin{small}
 \begin{tabular}{lrr}
\toprule
\textbf{Dataset} & \textbf{Identity} & \textbf{Zero-shot} \\
& \phantom{\textsc{All 3}} \\\addlinespace[1\aboverulesep]\addlinespace[1\cmidrulewidth]\addlinespace[1\belowrulesep]
\midrule
DE\textsubscript{A} & 30.16 & \textbf{40.94} \\
DE\textsubscript{R} & 43.57 & \textbf{55.92} \\
EN & \textbf{75.47} & 56.31 \\
ES & \textbf{72.29} & 64.39 \\
HU & 17.81 & \textbf{20.58} \\
IS & \textbf{47.77} & 42.95 \\
PT & 65.18 & \textbf{67.64} \\  
SL\textsubscript{B} & 39.84 & \textbf{50.21 }\\ 
SL\textsubscript{G} & \textbf{85.58} & 84.99 \\
SV & \textbf{59.24} & 50.65 \\
\midrule
Micro-Avg & 50.17&{\bf 52.96}\\
 \bottomrule
\end{tabular}
\end{small}
  \caption{Normalization accuracy on dev sets for zero-shot experiments.  Best results highlighted in bold.}
  \label{tab:results-zeroshot}
\end{table}

  \begin{figure*}[tb]
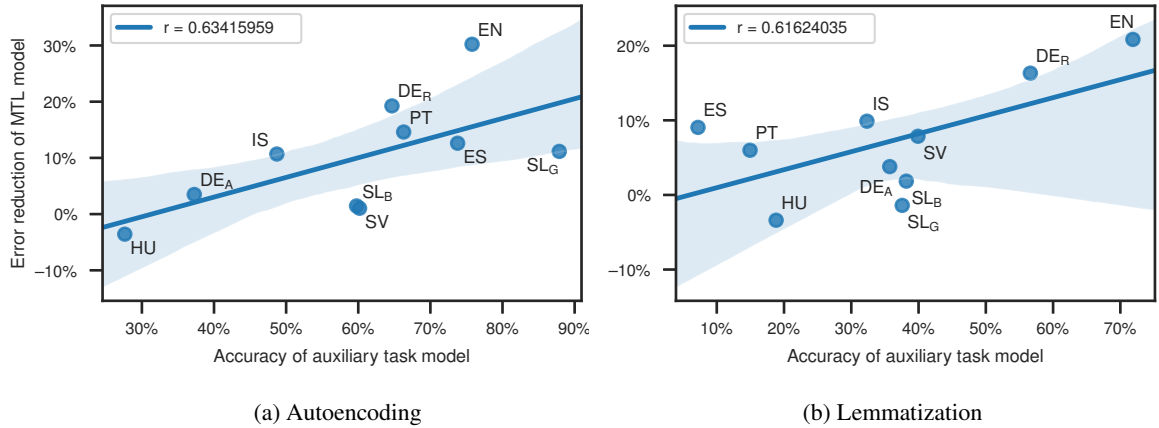

     \centering
     \begin{subfigure}[t]{0.45\linewidth}
	  \hspace{-2.8em}
 	  \input{plot_correlation_autoenc.pgf}
       \caption{Autoencoding}\label{fig:autoenc}
     \end{subfigure}
     \begin{subfigure}[t]{0.45\linewidth}
	  \hspace{-1.2em}
	  \input{plot_correlation_lemma.pgf}
       \caption{Lemmatization}\label{fig:lemma}
     \end{subfigure}
     \caption{Correlations (with 95\% confidence intervals) between the performance of an auxiliary task model applied to normalization data and the error reduction when using this task in a multi-task learning setup.}\label{fig:corr}
   \end{figure*}
   
   \begin{figure}[t]
   \centering
	  \hspace{-1.2em}
 	  \input{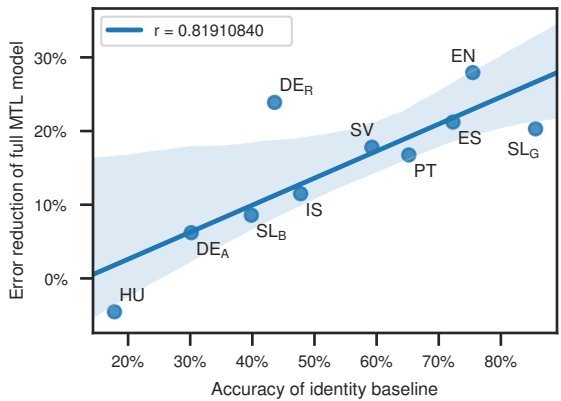}
       \caption{Correlation (with 95\% confidence interval) between the identity baseline and the error reduction of the full MTL~model with all three auxiliary tasks.}\label{fig:ident}
   \end{figure}

\paragraph{Results}
Table~\ref{tab:results-zeroshot} shows the accuracy of zero-shot normalization compared to the naive {\em identity baseline,} i.e., the accuracy obtained by simply leaving the input word forms unchanged.  The zero-shot approach improves over this baseline for half of the datasets, sometimes by up to 12~percentage points (DE\textsubscript{R}).  Micro-averaging the results shows an overall advantage for zero-shot learning.

\section{Analysis}

The experiment in Sec.\,\ref{sec:what-aux-task} has shown that not all auxiliary tasks are equally useful; furthermore, autoencoding is, on average, the most useful auxiliary task of the three, closely followed by lemmatization.  This gives rise to the hypothesis that MTL mostly helps the model learn the identity mappings between characters.

To analyze this, we feed the historical data into the {\em auxiliary models;} i.e., we treat them {\em as if} they were a historical text normalization model.  We then correlate their normalization accuracy with the error reduction over the baseline of the MTL~model using this auxiliary task.  Figure~\ref{fig:autoenc} shows a strong correlation for the autoencoding task, suggesting that the synergy between autoencoding and historical text normalization is higher {\em when the two tasks are very related}.  Figure~\ref{fig:lemma} shows the same correlation for lemmatization.

We can also compare the error reduction from MTL to the identity baseline (cf.\ Tab.\,\ref{tab:results-zeroshot}).  Figure~\ref{fig:ident} shows the correlation of these scores for the full MTL~model trained with all three auxiliary tasks.\footnote{The correlation is similar when using longest common subsequence or Levenshtein distance instead of accuracy.}  The strong correlation suggests that the regularization effect introduced by MTL is particularly helpful with tasks where there is a strong similarity between input and output; or, in other words, that {\em multi-task learning prevents the model from over-generalizing} based on the training data. 

\begin{figure*}
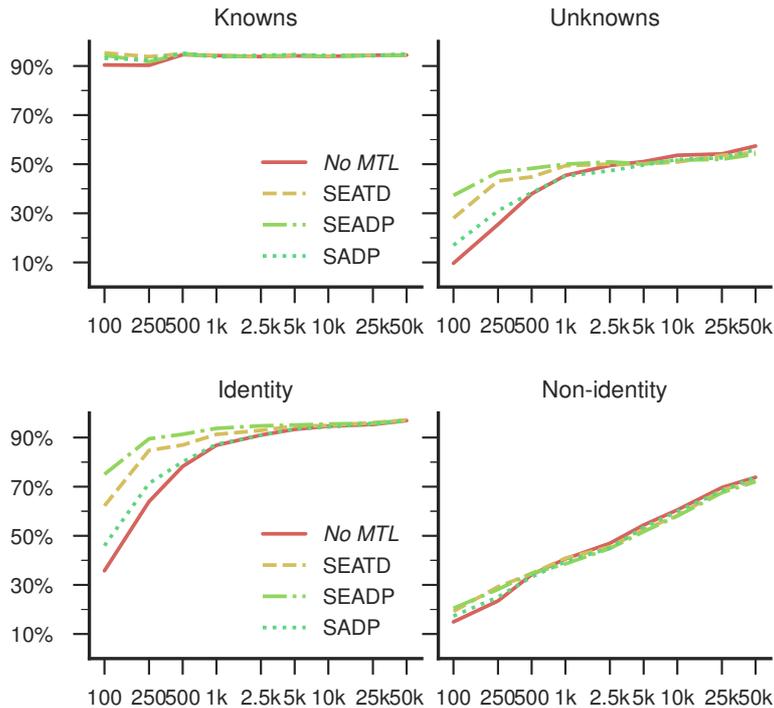

  \centering
     \input{plot_mtl_analysis_avg_unk.pgf}
     \input{plot_mtl_analysis_avg_id.pgf}
     \caption{Learning curves, micro-averaged over all datasets, for different subsets of the data.}
     \label{fig:id-unk-analysis}
\end{figure*}

The previous correlation scores only consider the performance of models trained on 1,000~tokens of historical data.  Sec.\,\ref{sec:learning-curves} showed that the benefit of MTL diminishes when the size of the historical training sets gets larger.  Figure~\ref{fig:id-unk-analysis} presents learning curves that have been micro-averaged over all ten datasets, but evaluated on different subsets of the data: (a)~tokens that have been seen during training (``knowns'') or not (``unknowns''); and (b)~tokens that stay identical in the reference normalization or not.  On average, the performance of the MTL~models is comparable to that of the single-task model for known tokens and non-identity normalizations.  In other words, most of the gain from MTL comes from helping the model learn the identity mappings, which becomes less relevant the more historical training data is available.

\section{Related work}
\label{sec:related-work}

On previous approaches to historical text normalization, \citet[Sec.\,2]{Bollmann2019} gives an extensive overview.
Common approaches include rule-based algorithms---with either manually crafted or automatically learned rules---or distance metrics to compare historical spellings to modern lexicon forms~\cite{Baron-Rayson2008,Bollmann2012,Pettersson-Megyesi-Nivre2013}.  Finite-state transducers are sometimes used to model this, but also to explicitly encode phonological transformations which often underlie the spelling variation~\cite{Porta-etal2013,Etxeberria-etal2016}.  

Character-based statistical machine translation~(CSMT) has been successfully applied to normalization on many languages~\cite{Pettersson-etal2013,Scherrer-Erjavec2016,Domingo-Casacuberta2018}; neural encoder--decoder models with character-level input can be seen as the neural equivalent to the statistical MT approach~\citep{Bollmann-etal2017,Tang-etal2018} and have been shown to be competitive with it~\citep{Robertson-Goldwater2018,Hamalainen-etal2018}, although \citet{Bollmann2019} suggests that they are still inferior to CSMT in low-resource scenarios.  

All these methods rely on individual word forms as their input; there is almost no work on incorporating sentence-level context for this task~\citep[but cf.][]{Jurish2010}.

\paragraph{MTL architectures}
In Sec.\,\ref{sec:what-to-share}, we explored \emph{what to share} between tasks in our multi-task architecture. A common approach is to share only the first layers~\citep[e.g.,][]{Yang-etal2016,Peng-Dredze2017}.  Multi-task encoder--decoder models will often keep the whole encoder and decoder task- or language-specific~\cite{Dong-etal2015,Luong-etal2016}.  \citet{Firat-etal2016} explore the effect of sharing the attentional component across all languages, while \citet{Anastasopoulos-Chiang2018} compare both parallel and cascading model configurations.

A different MTL~approach is to share \emph{all} parts of a model, but prepend a task-specific symbol to the input string to enable it to learn task-specific features (cf.\ Sec.\,\ref{sec:zeroshot}).  \citet{Milde-etal2017} use this approach for grapheme-to-phoneme conversion; \citet{Kann-etal2017} apply it to morphological paradigm completion.

\paragraph{Auxiliary tasks for MTL}
For \emph{which auxiliary task(s) to use} (Sec\,\ref{sec:what-aux-task}), few systematic studies exist.  Most approaches use tasks that are deemed to be related to the main task---e.g., combining machine translation with syntactic parsing~\citep{Kiperwasser-Ballesteros2018}---and justify their choice by the effectiveness of the resulting model. \citet{Bingel-etal2017} analyze beneficial task relations for MTL in more detail, but only consider sequence labelling tasks.
For zero-shot learning (Sec.\,\ref{sec:zeroshot}), we use an architecture very similar to \newcite{Johnson-etal2016}, also used for grapheme-to-phoneme mapping in \newcite{Peters:ea:17}.

\section{Conclusion}

We performed an extensive evaluation of a neural encoder--decoder model on historical text normalization, using little or even no training data for the target language, and using multi-task learning~(MTL) strategies to improve accuracy.  We found that sharing more components between main and auxiliary tasks is usually better, and autoencoding generally provides the most benefit for our task.  Analysis showed that this is mainly because MTL helps the model learn that most characters should stay the same, and that its beneficial effect vanishes as the size of the training set increases.  While our models did not beat the non-neural models of \citet{Bollmann2019}, we believe our work still provides interesting insights into the impact of MTL for low-resource scenarios.

\section*{Acknowledgments}

We would like to thank the anonymous reviewers of this as well as previous iterations of this paper for several helpful comments.

This research has received funding from the European Research Council under the ERC~Starting Grant LOWLANDS No.\ 313695.
Marcel Bollmann was partly funded from the European Union's Horizon~2020 research and innovation programme under the Marie Skłodowska-Curie grant agreement No.\ 845995.

\bibliography{phd}
\bibliographystyle{acl_natbib}

\end{document}